\documentclass[lettersize,journal]{IEEEtran}
\usepackage{amsmath,amsfonts}
\usepackage{algorithm}
\usepackage{algorithmicx}
\usepackage{array}
\usepackage[caption=false,font=normalsize,labelfont=sf,textfont=sf]{subfig}
\usepackage{textcomp}
\usepackage{stfloats}
\usepackage{url}
\usepackage{verbatim}
\usepackage[noend]{algpseudocode}  
 \usepackage[switch]{lineno}
\usepackage{graphicx}
\usepackage{multirow}
\usepackage[table,xcdraw]{xcolor}
\usepackage[graphicx]{realboxes}
\usepackage{color}
\usepackage{cite}
\usepackage{xcolor}
\usepackage[colorlinks,
            linkcolor=blue,       
            anchorcolor=blue,  
            citecolor=blue,        
            ]{hyperref}
\begin{document}

\title{Low-light Image Enhancement via CLIP-Fourier Guided Wavelet Diffusion}

\author{Minglong Xue, Jinhong He, Wenhai Wang and Mingliang Zhou 


}

%


\maketitle

\begin{abstract}
Low-light image enhancement techniques have significantly progressed, but unstable image quality recovery and unsatisfactory visual perception are still significant challenges. To solve these problems, we propose a novel and robust low-light image enhancement method via CLIP-Fourier Guided Wavelet Diffusion, abbreviated as CFWD. Specifically, CFWD leverages multimodal visual-language information in the frequency domain space created by multiple wavelet transforms to guide the enhancement process. Multi-scale supervision across different modalities facilitates the alignment of image features with semantic features during the wavelet diffusion process, effectively bridging the gap between degraded and normal domains. Moreover, to further promote the effective recovery of the image details, we combine the Fourier transform based on the wavelet transform and construct a Hybrid High Frequency Perception Module (HFPM) with a significant perception of the detailed features. This module avoids the diversity confusion of the wavelet diffusion process by guiding the fine-grained structure recovery of the enhancement results to achieve favourable metric and perceptually oriented enhancement. Extensive quantitative and qualitative experiments on publicly available real-world benchmarks show that our approach outperforms existing state-of-the-art methods, achieving significant progress in image quality and noise suppression. The project code is available at https://github.com/hejh8/CFWD.
\end{abstract}

\begin{IEEEkeywords}
Low-light image enhancement, diffusion model, multi-modal, Fourier transform, wavelet transform.
\end{IEEEkeywords}

\section{Introduction}
\IEEEPARstart{L}{OW}-Light image enhancement aims to enhance the quality and brightness of under-illuminated images. Due to the complex lighting conditions in the real world, relevant information in captured images is often lost through appropriate or significant masking. This poses a challenge to human visual perception and impedes the development and deployment of various downstream tasks, such as Target Detection \cite{dong2023retrieving}, Autonomous Driving \cite{li2021deep} and Text Detection \cite{xue2020arbitrarily}. Therefore, to address these challenges, low-light image enhancement techniques have been vigorously developed, and many related algorithms have been proposed. These techniques can be broadly categorized into traditional model-based approaches and data-driven deep learning-based approaches.

Traditional model-based low illumination image enhancement methods mainly construct physical models through methods such as histogram equalization \cite{pisano1998contrast} and Retinex theory \cite{land1971lightness}. Their focus is on using manually designed prior knowledge \cite{park2022histogram, fu2016weighted, li2018structure,sugimura2015enhancing} to optimize the degradation parameters of the image itself, and the effectiveness relies heavily on the accuracy of the manually created prior. However, low-illumination image enhancement is essentially a nonlinear problem with unknown degradation, so it is more difficult to use an artificial prior to adapt to various lighting conditions in an open scene.

With the development of deep learning, researchers have explored a large number of data-driven-based network learning methods \cite{zhang2023lrt, sun2015learning, guo2020zero, ren2019low, jiang2021enlightengan,zhang2023multi,wang2019low,lu2022progressive}. Wei et al. \cite{wei2018deep} constructed a deep-learning image decomposition algorithm based on the Retinex model. xu et al. \cite{xu2022snr} utilized a signal-to-noise ratio-aware transformer and a convolutional neural network (CNN) with spatially varying operations for restoration.  In addition, the recently emerged diffusion model \cite{song2019generative,song2020score} has attracted extensive attention from researchers in the field of image restoration \cite{ozdenizci2023restoring, saharia2022image, fei2023generative} due to its powerful generative and generalization capabilities. These methods essentially bridge the gap between the degraded and normal domains to obtain a clear normal image.

\begin{figure}[t]
        \centering
        \includegraphics[height=0.2\textwidth,width=0.47\textwidth]{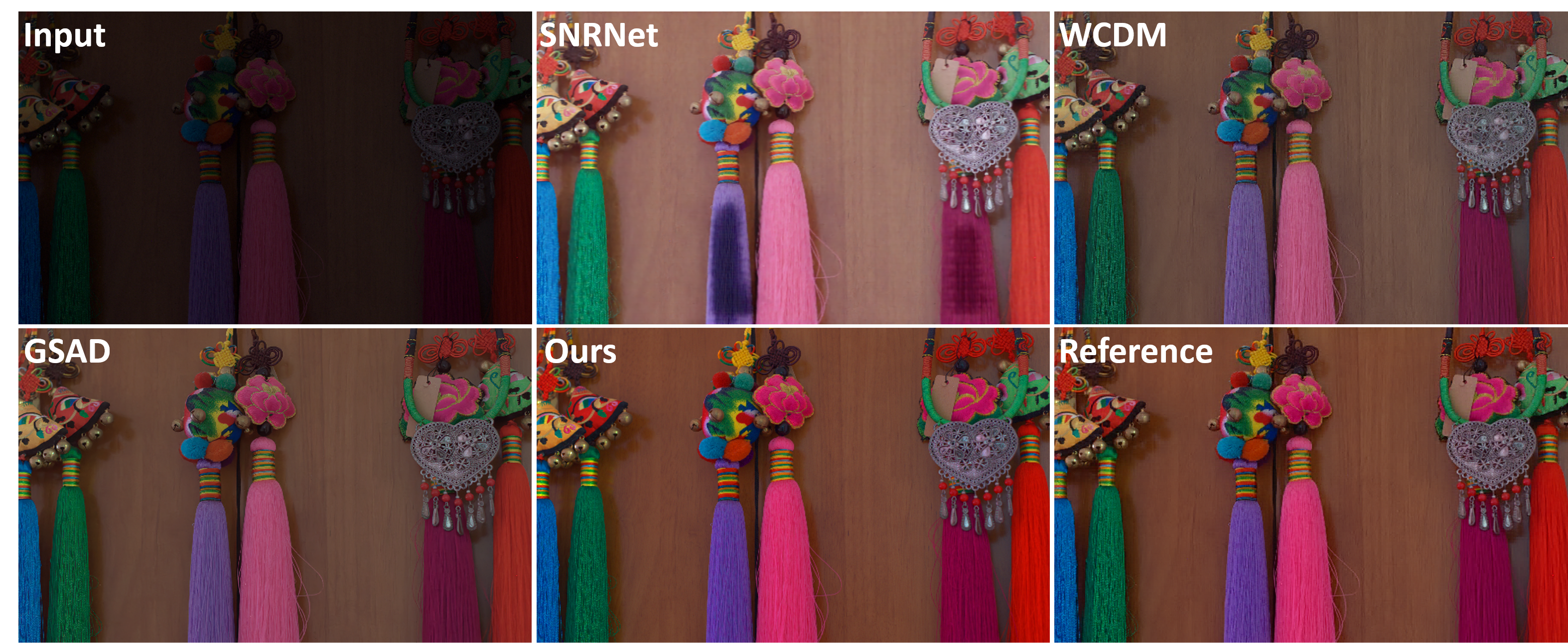}
        \caption{Visual comparison of our method with recent state-of-the-art methods. Other methods suffer from contrast degradation and noise artifacts. our method has the best visual perception.}
        \label{fig:1}
    \end{figure}

However, most existing methods such as GSAD and SNRNet tend to consider only supervising the enhancement process from the image level, neglecting the detailed reconstruction of the image and the role of multi-modal semantics in guiding the feature space. Such unimodal supervision produces suboptimal reconstruction of uncertain regions and poorer local structures, leading to the appearance of unsatisfactory visual results. For example, as shown in Fig. \ref{fig:1}, previous state-of-the-art approaches can suffer from color distortion, excessive noise, and redundant confusing information due to the lack of effective constraints and guidance. It is worth noting that diffusion models have diverse generative effects due to the stochastic nature of the inference process but also indirectly contribute to the difficulty of efficiently constraining noise and redundant information in image restoration tasks. 

Furthermore, for low-level visual tasks, the simple introduction of visual-language information does not reap significant performance. This may be due to the fact that image corruption creates difficulties for feature alignment, resulting in the inability of the visual-language model to capture the fine-grained gaps between degraded images and semantics effectively. Considering the above issues, our overall goal is to explore the introduction of multimodal semantics through frequency-domain diffusion iterations based on the Contrast-Language-Image-Pre-Training (CLIP) model to provide effective condition guidance and content constraints for the task of low-light image enhancement, and to achieve the enhancement of low-light image under different spatial illumination conditions.
%

Inspired by \cite{jiang2023low}, we adopt the wavelet diffusion model to establish a mapping between low-light and normal-light images, and also propose a novel CLIP and Fourier transform guided wavelet diffusion model (CFWD). Specifically, based on the pre-trained visual-language model CLIP, we gradually introduce semantic information in the frequency domain space of multiple wavelet transform decompositions, construct a multilevel semantic guidance network to alleviate the difficulty of multi-modal feature alignment, and impose multilevel conditional constraints on the diffusion process to achieve metric-friendly and perceptually oriented enhancement. In addition, we combine the wavelet transform and Fourier transform to construct a high-frequency hybrid space with significant perceptual capabilities. Appearance restoration of degraded images is explored from a spectral perspective, thus further avoiding the generative diversity of diffusion models. Extensive experimental results on public benchmark datasets show that CFWD significantly improves image quality assessment up to state-of-the-art while also providing better visualization. 

In summary, the contribution of this paper can be summarised as follows:
\begin{itemize}
\item We propose the method of CLIP-Fourier Guided Wavelet Diffusion (CFWD). This is the first successful introduction of multi-modal into the diffusion model-based low-light image enhancement work, which has a more realistic visual perception enhancement performance and a more stable generation effect.
\item To further enhance the conditional guidance, we designed a multi-level visual-language guidance network by combining frequency domain space and multi-modal for the first time. It effectively mitigates the multi-modal feature alignment problem caused by image corruption by gradually introducing visual-language information in the frequency domain in combination with the wavelet diffusion process. Meanwhile, the multilevel guidance of the enhancement process is achieved, which significantly improves the metric and visual perception. 
\item We construct high-frequency hybrid spaces with significant perceptual capabilities by exploring the effective combination of wavelet transform and Fourier transform. Effective constraints on the diversity of diffusion model generation are achieved, and the enhancement performance is effectively improved.
\end{itemize}

The remainder of this paper is structured as follows. In Section \ref{Related Works}, the related works are discussed. Section \ref{Preliminary} explains the conventional conditional diffusion model. In Section \ref{Method}, the proposed novel model method is described in detail. The relevant experimental setup and results are shown in Section \ref{EXPERIMENTS}. Section \ref{Conclusions} is the conclusion.

\section{Related Works} \label{Related Works}
\subsection{Traditional Approaches}
Low-light image enhancement has received extensive attention from researchers as an important support for various downstream tasks \cite{zhang2019rgb, dong2023retrieving, xue2020arbitrarily}. Traditional low-light image enhancement techniques mainly focus on constructing physical models using two types of methods, adaptive histogram equalisation\cite{pisano1998contrast} and Retinex theory\cite{land1971lightness}, which are processed by optimizing the parameter information of the image itself. The former class of algorithms optimizes pixel brightness based on the idea of histogram equalization, while the latter class of methods obtains the desired reflectance map (i.e., the normal image) by estimating the light from the low-light input and removing the effect of the estimated light. For example, \cite{wang2013naturalness} achieved enhancement of non-uniform images by balancing detail and naturalness through double logarithmic transformation. \cite{fu2016weighted} proposed a weighted variational model using regularisation terms to estimate the image illumination component and the reflection image. \cite{guo2016lime} used probing the maximum value in the RGB channel to estimate the illuminance of each pixel and subsequently enhanced the low-light image using a manually designed structural prior.
\subsection{Deep Learning Approaches}
The rapid development of deep learning has also triggered the enthusiasm of researchers to explore the field of low-light image enhancement. Numerous low-light enhancement algorithms through data-driven enhancement have been proposed one after another \cite{mao2016image,wang2020lightening,zhang2023multi,wang2019low,lamba2020harnessing}. Lore et al. \cite{lore2017llnet} proposed LLNet, the first network that applies deep learning to image enhancement, which is trained on degraded images through an encoder-decoder architecture. HDR-Net \cite{gharbi2017deep} combines deep networks with the ideas of bilateral grid processing and local affine color transformations with pairwise supervision. \cite{wei2018deep} proposed Retinex-Net, which first introduced Retinex theory to deep learning and constructed an end-to-end image decomposition algorithm. Zhang et al. \cite{zhang2019kindling} proposed the KinD method to improve the problem of producing unnatural enhancement results in Retinex-Net by introducing training loss and adjusting the network architecture. Enlightengan \cite{jiang2021enlightengan} used a generative inverse network as the main framework and was first trained using unpaired images. \cite{guo2020zero} constructed pixel level by stepwise derivation of the curve estimation convolutional neural network and designed a series of zero-reference training loss functions. \cite{xu2022snr} utilizes a signal-to-noise ratio aware transformer and a CNN model with spatially varying operations for recovery. Although all these methods have achieved remarkable results, they still face significant challenges in terms of generation quality and enhanced generalization performance due to the lack of effective supervision and efficient reconstruction of the content.

Furthermore, Efficient cross-modal learning has opened up new ideas for computer vision and has been greatly developed. Radford et al. \cite{radford2021learning} proposed to learn a priori knowledge from large-scale image-text data pairs in order to construct a visual language model CLIP for efficient image classification and task migration with zero-sample training. \cite{liang2023iterative} efficiently performed region enhancement on backlit images by iteratively learning the prompt text from a frozen pre-trained CLIP model. To the best of our knowledge, compared to other methods, we are the first to successfully introduce multi-modal learning in a diffusion model-based low-light image enhancement method and achieve significant performance improvements.

\begin{figure*}[t]
        \centering
        \includegraphics[height=0.3\textwidth,width=\textwidth]{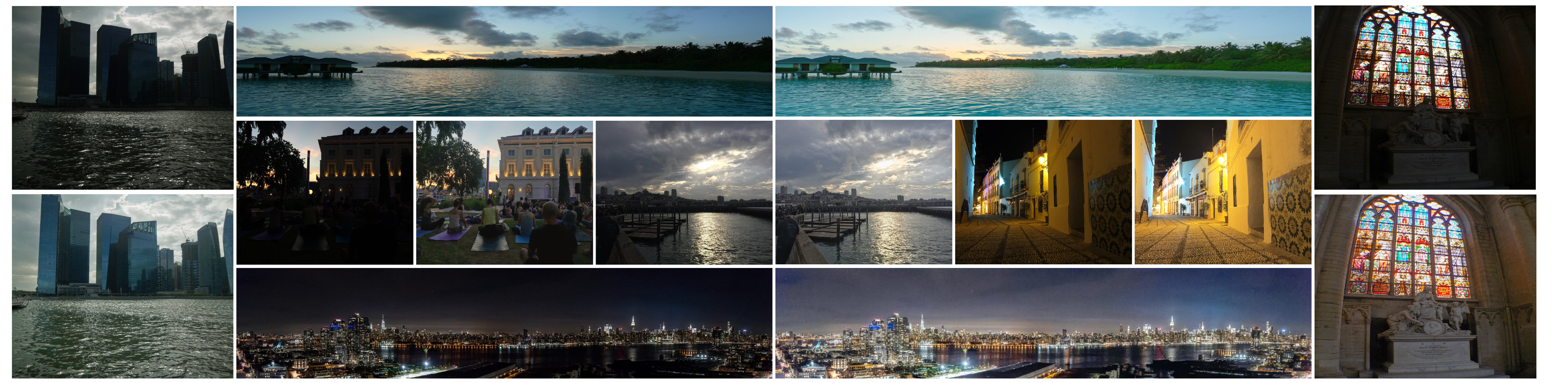}
        \caption{Representative visual examples by enhancing low-light images using CFWD. All of these images have either 2k resolution or 4k resolution.}
        \label{fig:2}
    \end{figure*}

\subsection{Diffusion Models Approaches}
Recently, diffusion-based generative models \cite{ho2020denoising} have achieved amazing results with the exploration of many researchers. Meanwhile, low-level visual tasks \cite{luo2023image, lugmayr2022repaint, ren2022image,yue2023dif,liao2022unsupervised} have also gained significant progress as a result. Saharia et al. \cite{saharia2022image} adopt a direct cascading approach, integrating low-resolution measurements and latent codes as inputs to train conditional diffusion models for restoration. WeatherDiff \cite{ozdenizci2023restoring} introduces a block-based diffusion model aimed at recuperating images taken in adverse weather conditions, employing guidance across overlapping blocks during the inference stage. 

Moreover, for low-light image enhancement, researchers have also recently favoured diffusion model-based approaches. Fei et al. \cite{fei2023generative} utilize the a priori knowledge embedded in a pre-trained diffusion model to address linear inverse problems effectively. Jiang et al. \cite{jiang2023low} advances a diffusion model rooted in wavelet transform tailored for enhancing images captured in low-light environments, achieving content stabilization through forward diffusion and denoising processes during training. \cite{hou2024global} introduced a diffusion model with a global structure-aware regularisation scheme for the enhancement of degraded images. Different from CFWD, the existing diffusion model approach does not allow for effective guidance and supervision during the enhancement process, leading to unnatural colours and numerous noises during inference. This seriously affects human visual perception and downstream task applications.


\section{Preliminary} \label{Preliminary}

Diffusion models \cite{ho2020denoising,song2020denoising} to train Markov chains by variational inference. It converts complex data into completely random data by adding noise and gradually predicts the noise to recover the expected clean image. Consequently, it usually includes the forward diffusion process and reverse inference process.

The forward diffusion process primarily relies on incrementally introducing Gaussian noise with a fixed variance $\{\beta_t\in(0, I)\}_{t=1}^T$ into the input distribution ${x}_0$ until the T time steps approximate purely noisy data. This process can be expressed as:
\vspace{-6pt}
\begin{equation}
q(x_1,\cdots,x_T  | x_0) =\prod_{t=1}^{T}q(x_t  |x_{t-1}),
\tag{1}
\end{equation}

\begin{equation}
q(x_t|x_{t-1})=N(x_t;\sqrt{1-\beta_t}x_{t-1},\beta_tI),
\tag{2}
\end{equation}
where $x_t$ and $\beta_t$ are the corrupted noise data and the predefined variance at time step $t$. Respectively, $N$ denotes a Gaussian distribution. Furthermore, each time step $x_t$ of the forward diffusion process can be obtained directly by computing $x_0$.

The reverse inference process is to recover the original data from Gaussian noise. In contrast to the forward diffusion process, The reverse inference process relies on optimising the noise predictor to iteratively remove the noise and recover the data until the randomly sampled noise $\hat{x}_T\sim N(0, I)$ becomes clean data $\hat{x}_0$. Formulated as:
\begin{equation} \label{eq3}
p_\theta(\hat{x}_0,\cdots,\hat{x}_{T-1}  | x_T)=\prod_{t=1}^{T} p_\theta(\hat{x}_{t-1}|\hat{x}_t),
\tag{3}
\end{equation}

\begin{equation}
p_\theta(\hat{x}_{t-1}|\hat{x}_t)=N(\hat{x}_{t-1};\mu _\theta(\hat{x}_t,t),\sigma ^2_tI),
\tag{4}
\end{equation}
where $\mu _\theta$ is the diffusion model noise predictor, which is mainly optimized by the editing and data synthesis functions and used as a way to learn the conditional denoising process, as follows:
\begin{equation}
\mu _\theta=\frac{1}{\sqrt{\alpha_t}}(\hat{x}_t-\frac{\beta_t}{\sqrt{1-\overline{\alpha}_t}}\epsilon_\theta(\hat{x}_t,t)),
\tag{5}
\end{equation}
where $\epsilon_\theta$ is a function approximator intended to predict $\epsilon$ from $\hat{x}_t$, $\alpha_t=1-\beta_t$, $\overline{\alpha}_t$=$\prod_{i=1}^{t}\alpha_i$.

\section{Method} \label{Method}
\begin{figure*}[t]
        \centering
        \includegraphics[height=0.53\textwidth,width=\textwidth]{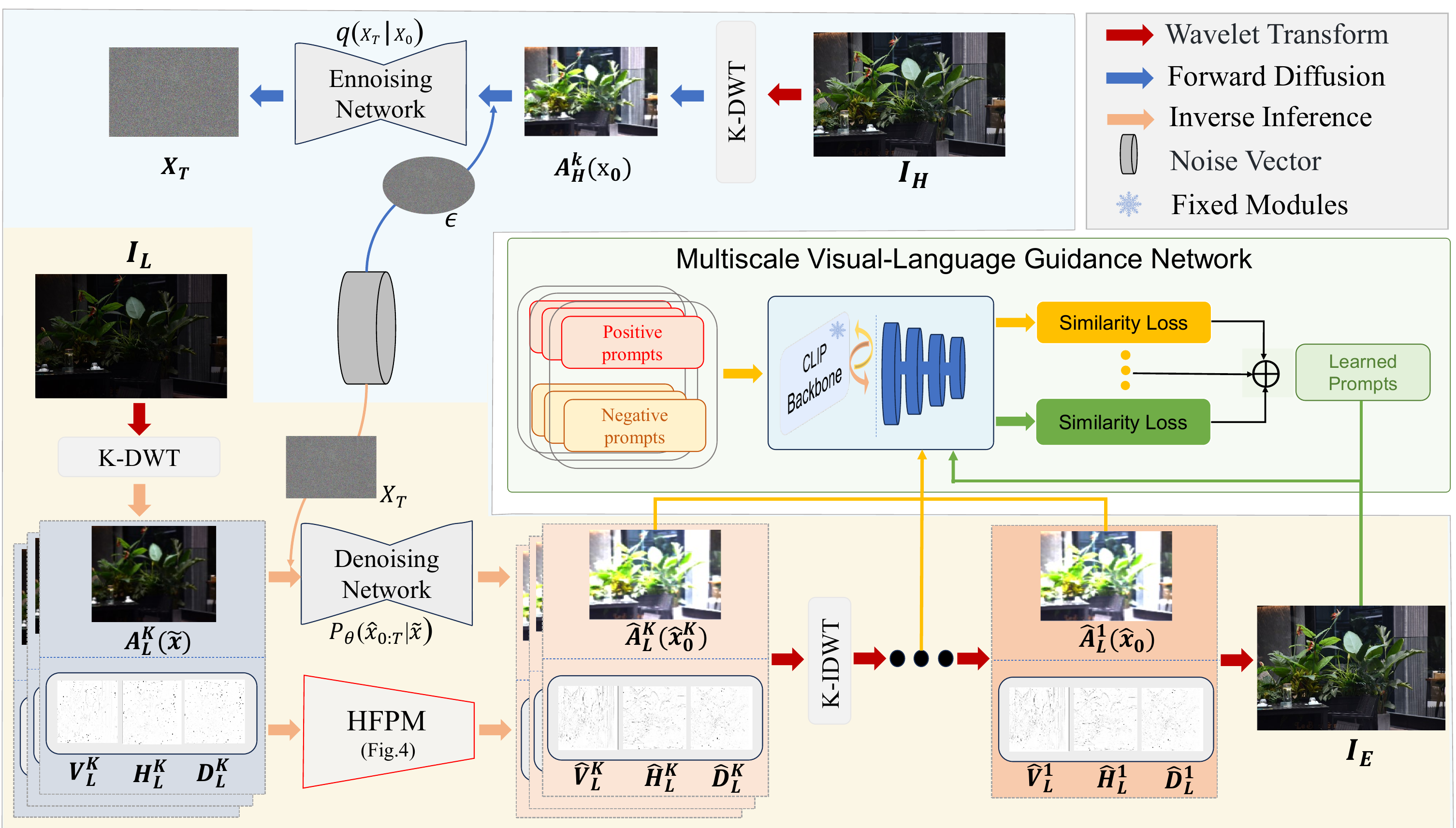}
        \caption{The overall workflow of our proposed CFWD. It first transforms the low-light input $I_L$ and normal image $I_H$ to the wavelet low-frequency domain $(A)$ for diffusion inference via the K-discrete wavelet transform (\rm{K-DWT}). We embed a multiscale visual guidance network to iteratively perform appearance guidance and content constraints by combining multiple wavelet domains in the inference process. In addition, the decomposed three high-frequency information $\{V_L, H_L, D_L\}$ we effectively augment by a high-frequency perception module (HFPM). Finally, the final enhancement result $I_E$ is obtained by inverse discrete wavelet transform (\rm K-IDWT).}
        \label{fig:3}
    \end{figure*}

As shown in Fig. \ref{fig:3}, inspired by \cite{jiang2023low}, our proposed method employs the wavelet diffusion model as a generative framework to reduce the consumption of computational resources. Meanwhile, we implement iterative guidance of the diffusion process to drive the appearance enhancement by effectively combining the visual-language and wavelet domains at multiple levels, which effectively mitigates the feature alignment difficulties of the visual-language model in the low-light image enhancement task. In addition, we explore the advantageous combination of wavelet transform and Fourier transform to construct a high-frequency perception module to guide the content reconstruction of diffusion models and bridge the gap between degraded and normal domains. Through the effective combination of multi-modal, frequency domain and diffusion models, we achieve high-quality visual enhancement effects and metric results. In this section, we first introduce the generative framework of this paper, i.e., the wavelet diffusion model, and then analyse in detail the multiscale visual-language guidance network and the high-frequency perception module.
\subsection{Wavelet Diffusion Model}

Existing diffusion models require high computational resources and are slow in efficiency. Therefore, we reduce the consumption of computational resources by transferring the diffusion process to the wavelet low-frequency domain via discrete wavelet transform. Specifically, in this part, the low-light image $I_L\in R^{H\times W\times C}$ and the normal image $I_H \in R^{H\times W\times C}$ are decomposed using the multiple discrete wavelet transform (K-DWT), where each time it is decomposed into four subbands:
\begin{equation}
\{A^K,V^K,H^K,D^K\}={\rm K\mbox{-}DWT}(I),
\tag{6}
\end{equation}
Where $A^K \in R^{\frac{H}{2^K}\times \frac{W}{2^K}\times C}$ denotes the low-frequency domain of the image after K-DWT. The $V^K, H^K$, and $D^K$ denote the high-frequency domain of the image in the vertical, horizontal, and diagonal directions, respectively.

Therefore, each discrete wavelet transform performed on an image is equivalent to downscaling its low-frequency domain to one-fourth of the original image. By shifting the diffusion process to take place in the wavelet low-frequency domain, we can significantly reduce the consumption of computational resources due to the substantial reduction in spatial dimensions.

Furthermore, we constrain the content diversity of the sampling process by performing forward diffusion in the wavelet low-frequency domain $A_H^K$ of the normal image $I_H$ and using the wavelet low-frequency domain $A_L^K$ of the degraded image $I_L$ as a conditional guide. Accordingly, Eq. \ref{eq3} can be rewritten as:
\begin{equation}
p_\theta(\hat{x}_{0:T}|\tilde{x})=p(\hat{x}_T)\prod_{t=1}^{T} p_\theta(\hat{x}_{t-1}|\hat{x}_t,\tilde{x}).
\tag{7}
\end{equation}

\begin{figure}[t]
\scalebox{0.5}{
        \includegraphics[height=0.54\textwidth,width=0.96\textwidth]{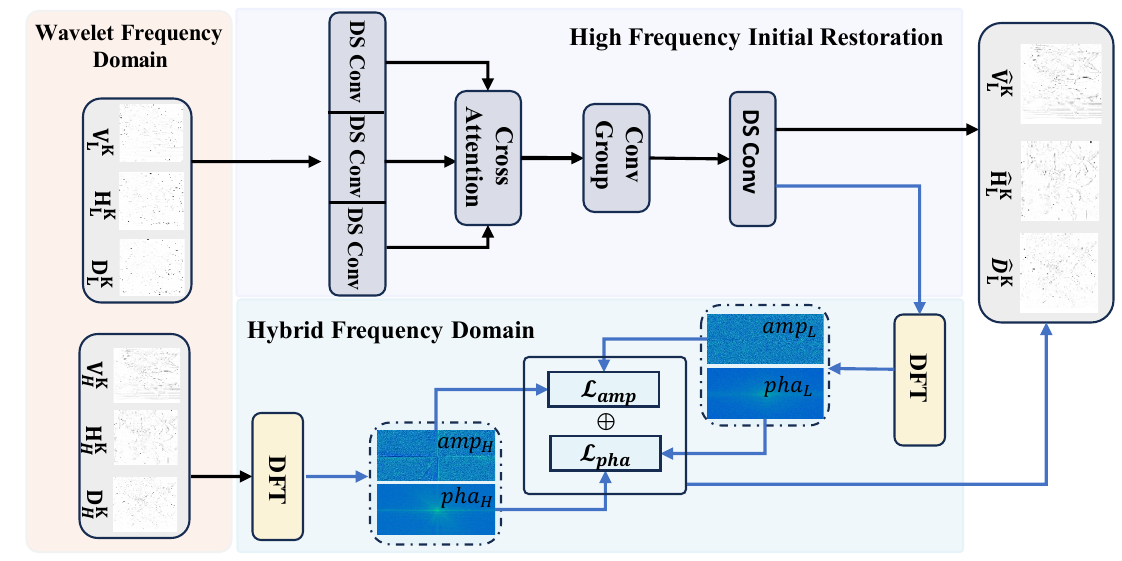}
        }
        \caption{Detailed architecture of our proposed High Frequency Perception Module (HFPM). DS Conv denotes depth-wise separable convolution, and DFT denotes Discrete Fourier Transform.}
        \label{fig:4}
    \end{figure}

\subsection{Multiscale visual-language Guidance Network}
Most of the existing low-light image enhancement algorithms reconstruct the appearance by image-level supervision through a single modality, which leads to difficulties in content reconstruction and significant degradation of the visual quality of the enhancement process. Meanwhile, simply applying visual-language models in low-level visual tasks does not obtain good performance. This may be due to their inability to capture fine-grained gaps in multi-modal semantics in degraded images, resulting in difficulties in aligning image features with text features.

\begin{figure}[t]
\scalebox{0.5}{
        \includegraphics[height=0.6\textwidth,width=0.96\textwidth]{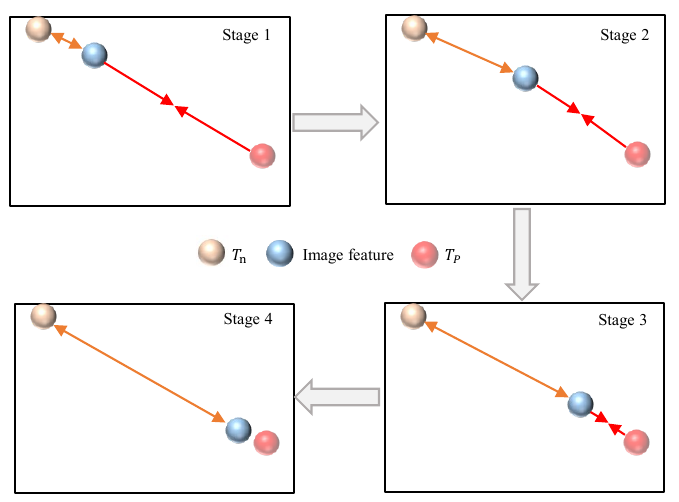}
        }
        \caption{The multiscale visual-language guidance network gradually promotes the alignment of image features with the positive prompts $T_p$ and continuously moves away from the negative prompts $T_n$. Stage 1 indicates without visual-language guidance.}
        \label{fig:5}
    \end{figure}

Therefore, we explore a combined frequency-domain diffusion and multi-modal approach to appearance guidance. The visual-language prompts are used in conjunction with the diffusion model to guide the appearance reconstruction of the wavelet domain of the image. Then, the enhancement results are used for multilevel semantic guidance to promote feature alignment between the image and the visual-language prompts, reaching a two-way iterative optimization effect. The image $A_L^K$ is first combined with visual-language prompts during the diffusion process, then performing coarse-grained feature alignment to obtain preliminary enhancement results $\hat{A}_L^K$. Using $\hat{A}_L^K$ after initially bridging the gap between the weak and normal light domains of $I_L$, we iteratively instruct its multiple wavelet low-frequency domains $\hat{A}_L^k(k\in[1, K-1])$ with the visual-language positive prompts $T_p$ and negative prompts $T_n$, expecting the low-light image to be enhanced in the direction of positive prompt $T_p$ and away from negative prompt $T_n$. As shown in Fig. \ref{fig:5}, when we set the wavelet transform scale $K=2$, through multi-scale semantic iterative guidance, the image is gradually enhanced in the desired direction. This further promotes the feature alignment between the image and the positive prompt $T_p$ and keeps moving away from the negative prompt $T_n$, realizing bidirectional appearance recovery.

We achieve alignment between images and prompt text features by freezing the latent space of the pre-trained visual-language model CLIP. By driving appearance recovery through visual-language prompts $\{T_p, T_n\}$, we significantly improve the contrast and illumination of the image and achieve stable sampling of the diffusion model. In addition, this section exploits cosine similarity to optimize network training, which can be formulated as follows:
\begin{equation}
	\begin{split}
\mathcal{L}_{Similarity\_1}&= \sum_{k=1}^{K}(\frac{\cos({\Phi}_{image} (\hat{A}_L^k ),\Phi_{text}(T_n )) }{\cos(\Phi_{image} (\hat{A}_L^k ),\Phi_{text}(T_p))}\\
&+\cos(\Phi_{image} (\hat{A}_L^{k}),\Phi_{text}(T_p))),
	\end{split}
 \tag{8}
\end{equation}
where $\Phi_{text}$ is the text encoder, and $\Phi_{image}$ is the image encoder. Along with the visual-language guidance, we use inverse discrete wavelet transform to recover the image until the final enhancement result $I_E$ is obtained. At the same time, we employ the learned prompts \cite{liang2023iterative} set to perform fine-grained multi-modal feature alignment on the final enhancement result, further expecting the enhancement result to reduce the distance from the target image, i.e.:
\begin{equation}
\mathcal{L}_{Similarity\_2}=\frac{e^{\cos(\Phi_{image} (I_E ),\Phi_{text}(T_n ))} }{\sum _{i\in\{T_p,T_n\}} e^{\cos(\Phi_{image} (I_E),\Phi_{text}(T_i))}}.
\tag{9}
\end{equation}

Thus, we can generalize the multiscale visual-language guidance loss as:
\begin{equation}
\mathcal{L}_{vlg}=\mathcal{L}_{Similarity\_1}+\mathcal{L}_{Similarity\_2}.
\tag{10}
\end{equation}

\subsection{High Frequency Perception Module}
Diffusion models have strong generative diversity, which becomes a limitation of algorithm performance for image enhancement and restoration tasks. Most of the current low-light image enhancement algorithms based on the diffusion model rely on image-level supervision with content reconstruction losses such as MSE and SSIM to achieve stable sampling of content. However, this does not provide significant content reconstruction of degraded images, which leads to content missing and visual degradation. Therefore, in order to further constrain the diffusion model, it is necessary to avoid generating content diversity and achieve visually oriented enhancement. Inspired by \cite{wang2023sgnet}, we explore the restoration of image high-frequency information from a frequency domain perspective.

The high-frequency perception module designed in this paper is shown in Fig. \ref{fig:4}. Compared with the low-frequency information, the high-frequency information generated by the discrete wavelet transform contains only the details and contours of the image, which can reduce the content interference for the Fourier transform and increase the ability to perceive the details of the image. Thus, we double-transform the image high frequency to construct the hybrid frequency domain space. We first perform detail enhancement \cite{jiang2023low} on the wavelet high-frequency information generated from the low-light image $I_L$ to obtain more contour structures and image parameters. Specifically, three high-frequency subbands $\{V_L^K, H_L^K, D_L^K\}$ are feature-extracted using depth-wise separable convolutions, and then the detail contours of $D$ are enhanced using $V, H$ combined with cross-attention. Subsequently, the enhanced three high-frequency subbands $\{\hat{V}_L^K,\hat{H}_L^K,\hat{D}_L^K\}$ are obtained by dilation convolutions \cite{hai2022combining} and depth-wise separable convolutions. After detail enhancement of the high-frequency information of $I_L$, we perform discrete Fourier transform $\rm DFT(\cdot)$ on $\{\hat{V}_L^K,\hat{H}_L^K,\hat{D}_L^K\}$ and $\{V_H^K,H_H^K,D_H^K\}$ obtained by the decomposition of the normal image $I_H$ to obtain the spectrum, i.e.:
\begin{equation}
amp_L,pha_L={\rm DFT}(\{\hat{V}_L^K,\hat{H}_L^K,\hat{D}_L^K\}),
\tag{11}
\end{equation}
\begin{equation}
amp_H,pha_H={\rm DFT}(\{{V}_H^K,{H}_H^K,{D}_H^K\}),
\tag{12}
\end{equation}
where $amp, pha$ denote the amplitude and phase of the image, respectively.

To further obtain an enhancement that is consistent with human perception, the method proposed in this paper employs the $L_1$ loss to minimize the information difference between the high-frequency information spectrograms of normal and low-light images:
\begin{equation}
\mathcal{L}_{spectral}=\vartheta_1\mathcal{L}_{amp}+\vartheta_2\mathcal{L}_{pha},
\tag{13}
\end{equation}

\begin{equation}
\mathcal{L}_{amp}=\frac{1}{K}\sum_{i=1}^{K}\parallel{amp_L^i-amp_H^i\parallel}_1,
\tag{14}
\end{equation}

\begin{equation}
\mathcal{L}_{pha}=\frac{1}{K}\sum_{i=1}^{K}\parallel{pha_L^i-pha_H^i\parallel}_1,
\tag{15}
\end{equation}
where $\vartheta_1$ and $\vartheta_2$ are the weighting parameters for the amplitude and phase losses, and $i$ is the scale of the current wavelet transform.

\subsection{Model Training}
In CFWD, the loss function can be divided into three main parts: diffusion loss, multi-scale semantic guided loss and content reconstruction loss. Among them, diffusion loss is used to optimize the noise prediction of the diffusion model. In order to initially constrain the content diversity, this paper shifts the diffusion process to the wavelet low-frequency domain to carry out and minimize their L2 distances. Accordingly, the objective function is denoted as:
\begin{equation}
\begin{split}
 \mathcal{L}_{diff} &= E_{t\sim[1,T]}E_{x_0\sim p(x_0)}E_{z_t\sim N(0,I)}\\ 
 &\parallel \epsilon_t-\epsilon_\theta(x_t,\tilde{x},t)\parallel^2 + \vert\vert\hat{A}_L^K-A_H^K \vert\vert^2.
\end{split}
\tag{16} 
\end{equation}

For content reconstruction loss, in addition to optimizing the spectral loss of details, we perform content reconstruction by combining MSE loss and SSIM loss to minimize the content difference between the recovered image $I_L$ content  and the reference image $I_H$ content, i.e. :
\begin{equation}
\begin{split}
    \mathcal{L}_{content}&=\sum_{l=0}^{4}{\gamma_l\parallel\Phi_{image}^l(I_E)-\Phi_{image}^l(I_H)\parallel ^2}\\
&+(1-SSIM(I_E,I_H)),
\end{split}
\tag{17}
\end{equation}
where $\gamma_l$ is the weight of layer $l$ of the image encoder in the ResNet101 CLIP model. 

Accordingly, by combining multiple losses, we significantly enhance the model performance and obtain a satisfactory visual perception, with the total loss denoted as:
\begin{equation}
\mathcal{L}_{total}=\mathcal{L}_{diff}+\mathcal{L}_{vlg}+\mathcal{L}_{spectral}+\mathcal{L}_{content}.
\tag{18}
\end{equation}

\begin{table*}[t]
\renewcommand\arraystretch{1.4}
\caption{Quantitative evaluation of different methods on LOLv1 \cite{wei2018deep} , LOLv2-Real\_captured \cite{yang2021sparse} , and LSRW  datasets \cite{hai2023r2rnet}. The best and second performance are marked in {\textcolor[HTML]{FF0000}{red}} and {\color[HTML]{1A16CE}{blue}}, respectively.}
\scalebox{0.90}{
\begin{tabular}{l|c|cccc|cccc|cccc}
\hline
                        &                             & \multicolumn{4}{c|}{LOLv1}                                                                                                  & \multicolumn{4}{c|}{LOLv2-Real\_captured}                                                                                   & \multicolumn{4}{c}{LSRW}                                                                                                    \\ \cline{3-14} 
\multirow{-2}{*}{Methods} & \multirow{-2}{*}{Reference} & PSNR↑                          & SSIM↑                         & LPIPS↓                        & FID↓                           & PSNR↑                          & SSIM↑                        & LPIPS↓                        & FID↓                           & PNSR↑                          & SSIM↑                        & LPIPS↓                        & FID↓                           \\ \hline
RetinexNet              & BMVC'18                     & 26.316                        & 0.844                        & 0.219                        & 48.037                        & 17.715                        & 0.652                        & 0.436                        & 133.905                       & 15.609                        & 0.414                        & 0.454                        & 108.350                       \\
DSLR                    & TMM'20                      & 14.816                        & 0.572                        & 0.375                        & 104.428                       & 17.000                        & 0.596                        & 0.408                        & 114.306                       & 15.259                        & 0.441                        & 0.464                        & 84.930                        \\
DRBN                    & CVPR'20                     & 16.774                        & 0.462                        & 0.417                        & 126.266                       & 18.466                        & 0.768                        & 0.352                        & 89.085                        & 16.734                        & 0.507                        & 0.457                        & 80.727                        \\
Zero-DCE                & CVPR'20                     & 14.861                        & 0.559                        & 0.385                        & 87.270                        & 18.194                        & 0.649                        & 0.390                        & 84.123                        & 15.858                        & 0.454                        & 0.421                        & 65.690                        \\
MIRNet                  & ECCV'20                     & 24.138                        & 0.830                        & 0.250                        & 69.179                        & 20.020                        & 0.820                        & 0.233                        & 49.108                        & 16.470                        & 0.477                        & 0.430                        & 93.811                        \\
Zero-DCE++              & TPAMI'21                    & 14.682                        & 0.472                        & 0.407                        & 87.552                        & 17.461                        & 0.490                        & 0.427                        & 81.727                        & 16.210                        & 0.457                        & 0.431                        & 59.959                        \\
EnlightenGAN            & TIP'21                      & 17.483                        & 0.651                        & 0.390                        & 95.028                        & 18.676                        & 0.678                        & 0.364                        & 84.044                        & 17.081                        & 0.470                        & 0.420                        & 69.184                        \\
ReLLIE                  & ACM MM'21                   & 11.437                        & 0.482                        & 0.375                        & 95.510                        & 14.400                        & 0.536                        & 0.334                        & 79.838                        & 13.685                        & 0.422                        & 0.404                        & 65.221                        \\
RUAS                    & CVPR'21                     & 16.405                        & 0.499                        & 0.382                        & 102.013                       & 15.351                        & 0.495                        & 0.395                        & 94.162                        & 14.271                        & 0.461                        & 0.501                        & 78.392                        \\
DDIM                    & ICLR'21                     & 16.521                        & 0.776                        & 0.376                        & 84.071                        & 15.280                        & 0.788                        & 0.387                        & 76.387                        & 14.858                        & 0.486                        & 0.495                        & 71.812                        \\
CDEF                    & TMM'22                      & 16.335                        & 0.585                        & 0.407                        & 90.620                        & 19.757                        & 0.630                        & 0.349                        & 74.055                        & 16.758                        & 0.465                        & 0.399                        & 62.780                        \\
SCI                     & CVPR'22                     & 14.784                        & 0.526                        & 0.392                        & 84.907                        & 17.304                        & 0.540                        & 0.345                        & 67.624                        & 15.242                        & 0.419                        & 0.404                        & 56.261                        \\
URetinex-Net            & CVPR'22                     & 19.842                        & 0.824                        & 0.237                        & 52.383                        & 21.093                        & 0.858                        & 0.208                        & 49.836                        & 18.271                        & 0.518                        & 0.419                        & 66.871                        \\
SNRNet                  & CVPR'22                     & 24.609                        & 0.841                        & 0.262                        & 56.467                        & 21.480                        & 0.849                        & 0.237                        & 54.532                        & 16.499                        & 0.505                        & 0.419                        & 65.807                        \\
Uformer                 & CVPR'22                     & 19.001                        & 0.741                        & 0.354                        & 109.351                       & 18.442                        & 0.759                        & 0.347                        & 98.138                        & 16.591                        & 0.494                        & 0.435                        & 82.299                        \\
Restormer               & CVPR'22                     & 20.614                        & 0.797                        & 0.288                        & 72.998                        & 24.910                        & 0.851                        & 0.264                        & 58.649                        & 16.303                        & 0.453                        & 0.427                        & 69.219                        \\
Palette                 & SIGGRAPH'22                 & 11.771                        & 0.561                        & 0.498                        & 108.291                       & 14.703                        & 0.692                        & 0.333                        & 83.942                        & 13.570                        & 0.476                        & 0.479                        & 73.841                        \\
UHDFour                 & ICLR'23                     & 23.093                        & 0.821                        & 0.259                        & 56.912                        & 21.785                        & 0.854                        & 0.292                        & 60.837                        & 17.300                        & 0.529                        & 0.443                        & 62.032                        \\
CLIP-LIT                & ICCV'23                     & 12.394                        & 0.493                        & 0.397                        & 108.739                       & 15.262                        & 0.601                        & 0.398                        & 100.459                       & 13.483                        & 0.405                        & 0.425                        & 77.065                        \\
NeRCo                   & ICCV'23                     & 22.946                        & 0.785                        & 0.311                        & 76.727                        & 25.172                        & 0.785                        & 0.338                        & 84.534                        & {\color[HTML]{1A16CE} 19.456} & 0.549                        & 0.423                        & 64.555                        \\
WeatherDiff             & TPAMI'23                    & 17.913                        & 0.811                        & 0.272                        & 73.903                        & 20.009                        & 0.829                        & 0.253                        & 59.670                        & 16.507                        & 0.487                        & 0.431                        & 96.050                        \\
GDP                     & CVPR'23                     & 15.904                        & 0.540                        & 0.431                        & 112.363                       & 14.290                        & 0.493                        & 0.435                        & 102.416                       & 12.887                        & 0.362                        & 0.412                        & 76.908                        \\
GSAD                    & NeurIPS'23                  & {\color[HTML]{1A16CE} 27.629} & {\color[HTML]{FF0000} 0.876} & {\color[HTML]{FF0000} 0.188} & {\color[HTML]{1A16CE} 43.659} & 28.805                        & {\color[HTML]{FF0000} 0.894} & {\color[HTML]{1A16CE} 0.201} & {\color[HTML]{1A16CE} 41.456} & 19.418                        & 0.542                        & 0.386                        & 57.219                        \\
WCDM                    & TOG'23                      & 26.316                        & 0.844                        & 0.219                        & 48.037                        & {\color[HTML]{1A16CE} 28.875} & 0.874                        & 0.203                        & 45.395                        & 19.281                        & {\color[HTML]{1A16CE} 0.552} & {\color[HTML]{FF0000} 0.350} & {\color[HTML]{FF0000} 45.294} \\ \hline
CFWD(Ours)              & -                           & {\color[HTML]{FF0000} 29.185} & {\color[HTML]{1A16CE} 0.872} & {\color[HTML]{1A16CE} 0.197} & {\color[HTML]{FF0000} 40.987} & {\color[HTML]{FF0000} 29.855} & {\color[HTML]{1A16CE} 0.891} & {\color[HTML]{FF0000} 0.193} & {\color[HTML]{FF0000} 34.814} & {\color[HTML]{FF0000} 19.566} & {\color[HTML]{FF0000} 0.572} & {\color[HTML]{1A16CE} 0.374} & {\color[HTML]{1A16CE} 47.606} \\ \hline
\end{tabular}
}
\label{tab:1}
\end{table*}

\section{EXPERIMENTS} \label{EXPERIMENTS}
\subsection{Experimental Settings}
{\bf Dataset.} Our network is trained and evaluated on the LOLv1 dataset \cite{wei2018deep}, which contains 500 real-world low/normal light image pairs, of which 485 image pairs are used for training, and 15 image pairs are used for evaluation. In addition, we employ two other real-world pairwise datasets, LOLv2-Real\_captured \cite{yang2021sparse}, and LSRW \cite{hai2023r2rnet}, to evaluate the performance of our proposed network. Specifically, the LOLv2-Real\_captured dataset contains 689 low/normal light image pairs for training and 100 for testing. Most low-light images were collected by varying the exposure time and ISO and fixing other camera parameters. The LSRW dataset contains 5,650 image pairs captured in a variety of scenarios. 5,600 image pairs were randomly selected as the training set, and the remaining 50 pairs were used for evaluation. To evaluate the generalization ability of the proposed method in this paper, we tested our method on the BAID \cite{lv2022backlitnet} test dataset, which consists of 368 backlit images with 2K resolution. In addition, we also tested on two unpaired datasets, LIME \cite{guo2016lime} and DICM \cite{lee2013contrast}.

{\bf Implementation Details.} We implemented our method with PyTorch on two NVIDIA RTX 3090 GPUs. The network was set up with a total of $2\times10^5$ iterations, using the Adam optimizer, with the initial learning rate set to $1\times10^{-4}$, and the batch size and patch size set to $16$ and $256\times256$, respectively. 

{\bf Evaluation Metrics.} For the real-world paired datasets we tested, we used two full-reference distortion measures, PNSR and SSIM \cite{wang2004image}, as well as two perceptual metrics, LPIPS \cite{zhang2018unreasonable} and FID \cite{heusel2017gans}, to evaluate the performance and visual satisfaction of our approach. Higher PSNR or SSIM implies more realistic restoration results, while lower LPIPS or FID indicates higher quality details, brightness and hue. In addition, for the unpaired datasets LIME and DICM, we used three non-reference perceptual metrics: NIQE \cite{mittal2012making}, BRISQUE \cite{mittal2012no}, and PI \cite{blau20182018} to evaluate the visual quality of the enhancement results. The lower the metrics, the better the visual quality.


{\bf Comparison Methods}. To verify the effectiveness of the method proposed in this paper, we compared it with the State-of-the-art methods in recent years, including RetinexNet \cite{wei2018deep}, DSLR \cite{lim2020dslr}, DRBN \cite{yang2020fidelity}, Zero-DCE \cite{guo2020zero}, Zero-DCE++\cite{li2021learning}, MIRNet \cite{zamir2020learning}, EnlightenGAN \cite{jiang2021enlightengan}, ReLLIE \cite{zhang2021rellie}, RUAS \cite{liu2021retinex}, DDIM \cite{song2020denoising}, SCI \cite{ma2022toward}, URetinex-Net \cite{wu2022uretinex}, SNRNet \cite{xu2022snr}, Palette \cite{saharia2022palette}, Uformer \cite{wang2022uformer}, Restormer \cite{zamir2022restormer}, CDEF \cite{lei2022low}, UHDFour \cite{li2023embedding}, CLIP-LIT \cite{liang2023iterative}, NeRCo\cite{yang2023implicit}, WeatherDiff \cite{ozdenizci2023restoring}, GDP \cite{fei2023generative}, WCDM \cite{jiang2023low} and GSAD \cite{hou2024global}.

\subsection{Results}

\begin{figure*}[t]
        \centering
        \includegraphics[height=0.72\textwidth,width=\textwidth]{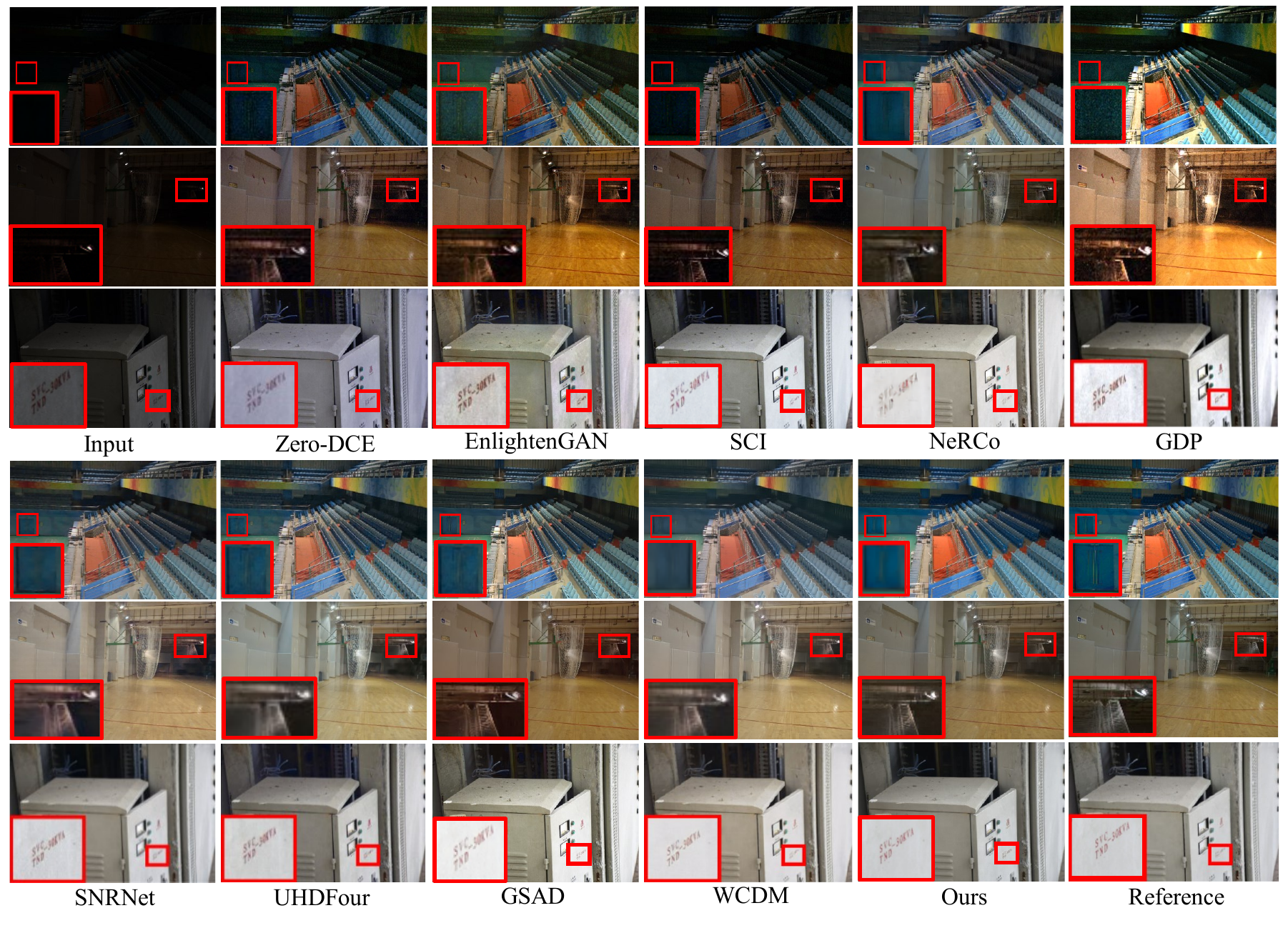}  
        \caption{Visual comparison of our method with State-of-the-art methods on LOLv1 \cite{wei2018deep}(row 1), LOLv2-Real\_captured \cite{yang2021sparse}(row 2), and LSRW \cite{hai2023r2rnet}(row 3) datasets from various years in recent years. Our method is closer to a normal image, best viewed by zooming in. }
        \label{fig:6}
    \end{figure*}

\begin{table}[t]
\renewcommand\arraystretch{1.3}
\caption{Quantitative comparison of 2K resolution backlight images from the BAID \cite{lv2022backlitnet} dataset. }
\scalebox{1.2}{
\begin{tabular}{l|cccc}
\hline
Methods      & PSNR↑                                                  & SSIM↑                         & LPIPS↓                         & FID↓                            \\ \hline
Zero-DCE++   & 16.021                                                & 0.832                        & 0.240                        & 47.030                        \\
EnlightenGAN & 17.957                                                & 0.866                        & {\color[HTML]{3531FF} 0.125} & 47.045                        \\
SCI          & 16.639                                                & 0.768                        & 0.197                        & 41.458                        \\
SNRNet       & 17.325                                                & 0.754                        & 0.398                        & 63.463                        \\
CLIP-LIT     & 21.611                                                & 0.883                        & 0.159                        & 27.926                        \\
UHDFour      & 18.541                                                & 0.713                        & 0.319                        & 36.025                        \\
WCDM         & {\color[HTML]{3531FF} 26.042}                         & {\color[HTML]{3531FF} 0.915} & {\color[HTML]{000000} 0.130} & {\color[HTML]{3531FF} 15.870} \\ \hline
CFWD(Ours)   & \cellcolor[HTML]{FFFFFF}{\color[HTML]{FE0000} 26.918} & {\color[HTML]{FE0000} 0.917} & {\color[HTML]{FE0000} 0.118} & {\color[HTML]{FE0000} 14.852} \\ \hline
\end{tabular}
}
\label{tab:2}
\vspace{-10pt}
\end{table}

\begin{table}[h]
\renewcommand\arraystretch{1.4}
\caption{Quantitative comparison on LIME \cite{guo2016lime} and DICM \cite{lee2013contrast} datasets. Our method performs the best consistently.}
\scalebox{0.85}{
\begin{tabular}{l|ccc|ccc}
\hline
                          & \multicolumn{3}{c|}{DICM}                                                                   & \multicolumn{3}{c}{LIME}                                                                    \\ \cline{2-7} 
\multirow{-2}{*}{Metheds} & NIQE↓                          & BRISQUE↓                        & PI↓                            & NIQE↓                          & BRISQUE↓                        & PI↓                            \\ \hline
DRBN                      & 4.369                        & 30.708                        & 3.782                        & 4.562                        & 29.564                        & 3.573                        \\
Zero-DCE                  & 3.414                        & 36.452                        & 2.911                        & 3.771                        & 18.481                        & 2.759                        \\
MIRNet                    & 4.021                        & 22.104                        & 3.391                        & 4.378                        & 28.623                        & 2.998                        \\
RUAS                      & 5.119                        & 41.897                        & 4.127                        & 4.702                        & 29.601                        & 3.479                        \\
DDIM                      & 3.899                        & 19.787                        & 3.013                        & 3.899                        & 24.474                        & 3.059                        \\
EnlightenGAN              & 3.439                        & 14.175                        & 2.719                        & 3.656                        & 14.854                        & 2.832                        \\
SCI                       & 3.519                        & 25.289                        & 2.824                        & 4.163                        & 17.094                        & 2.908                        \\
URetinex-Net              & 4.774                        & 24.544                        & 3.565                        & 4.694                        & 29.022                        & 3.313                        \\
SNRNet                    & 4.070                        & 26.179                        & 3.926                        & 5.691                        & 34.187                        & 4.636                        \\
CLIP-Lit                  & 3.557                        & 26.991                        & 2.589                        & 3.989                        & 19.422                        & 2.813                        \\
NeRCo                     & {\color[HTML]{3531FF} 3.329} & 19.586                        & 2.890                        & 3.803                        & 21.164                        & 2.888                        \\
UHDFour                   & 4.231                        & {\color[HTML]{3531FF} 13.174} & 3.186                        & 4.627                        & 15.930                        & 3.344                        \\
GDP                       & 4.358                        & 19.294                        & 2.887                        & 4.186                        & 22.022                        & 3.109                        \\
GSAD                      & 3.735                        & 20.296                        & 2.894                        & 4.578                        & 26.356                        & 3.492                        \\
WCDM                      & 3.364                        & 15.862                        & {\color[HTML]{FE0000} 2.364} & {\color[HTML]{3531FF} 3.597} & {\color[HTML]{3531FF} 14.474} & {\color[HTML]{FE0000} 2.605} \\ \hline
CFWD(Ours)                & {\color[HTML]{FE0000} 3.322} & {\color[HTML]{FE0000} 10.955} & {\color[HTML]{3531FF} 2.699} & {\color[HTML]{FE0000} 3.568} & {\color[HTML]{FE0000} 10.141} & {\color[HTML]{3531FF} 2.686} \\ \hline
\end{tabular}
}
\label{tab:3}
\vspace{-10pt}
\end{table}

{\bf Quantitative Comparison.} Firstly, we compare our method with all state-of-the-art methods on the LOLv1 \cite{wei2018deep}, LOLv2-Real\_captured \cite{yang2021sparse}, and LSRW \cite{hai2023r2rnet} test sets. As shown in Table \ref{tab:1}, our method achieves state-of-the-art quantization performance in several metrics compared to all methods. In particular, the significant improvements in PSNR and FID provide compelling evidence for the superior perceived quality of our method. Specifically, for two distortion metrics, our method obtains all firsts in PSNR evaluation, achieving performance improvements of 1.556dB, 0.98dB, and 0.11dB in the LOLv1, LOLv2-Real\_captured, and LSRW datasets, respectively. Furthermore, our method achieves the second-best SSIM quantisation performance on the LOLv1 and LOLv2-Real\_captured datasets. Compared to the third-place WCDM, our method has a significant improvement of 0.028 (LOLv1) and 0.017 (LOLv2-Real\_captured ), respectively, while for the first-place GSAD, we only have a small difference of 0.004 and 0.003. For two perceptual metrics (i.e., LPIPS and FID), our method meets the quantitative criteria on the LOLv2-Real\_captured dataset and is well ahead of competing methods. We are also significantly competitive on the LOLv1 and LSRW datasets, obtaining three second-place as well as one first-place quantitative performances. This indicates that the method proposed in this paper can generate recovered images with satisfactory visual quality, further demonstrating the effectiveness of our method. Table \ref{tab:2} also provides a quantitative comparison of some state-of-the-art methods on the BAID \cite{lv2022backlitnet} test dataset. From the evaluation metrics, our method outperforms all the state-of-the-art methods, which indicates that our proposed method is more effective in terms of generalisation ability and high-resolution low-light image restoration.

\begin{figure*}[t]
        \centering
        \includegraphics[height=0.31\textwidth,width=\textwidth]{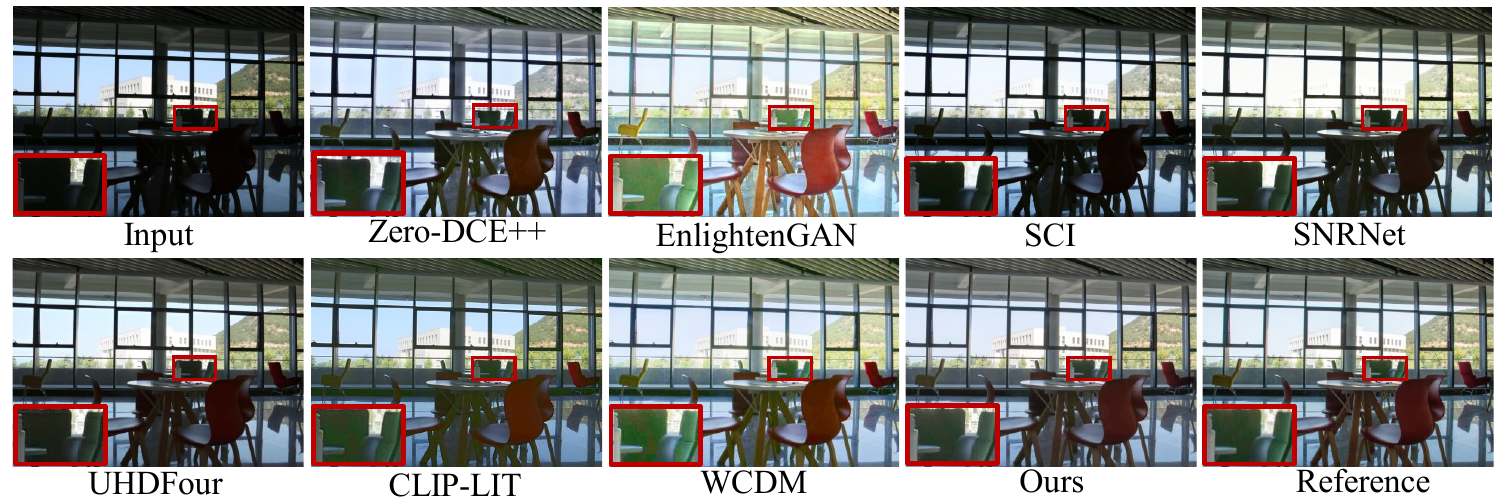}
        \caption{Visual comparison of 2K resolution backlight images of our method and competing methods on BAID \cite{lv2022backlitnet} test set. It is best viewed by zooming in. }
        \label{fig:7}
    \end{figure*}

\begin{figure*}[t]
        \centering
        \includegraphics[height=0.7\textwidth,width=\textwidth]{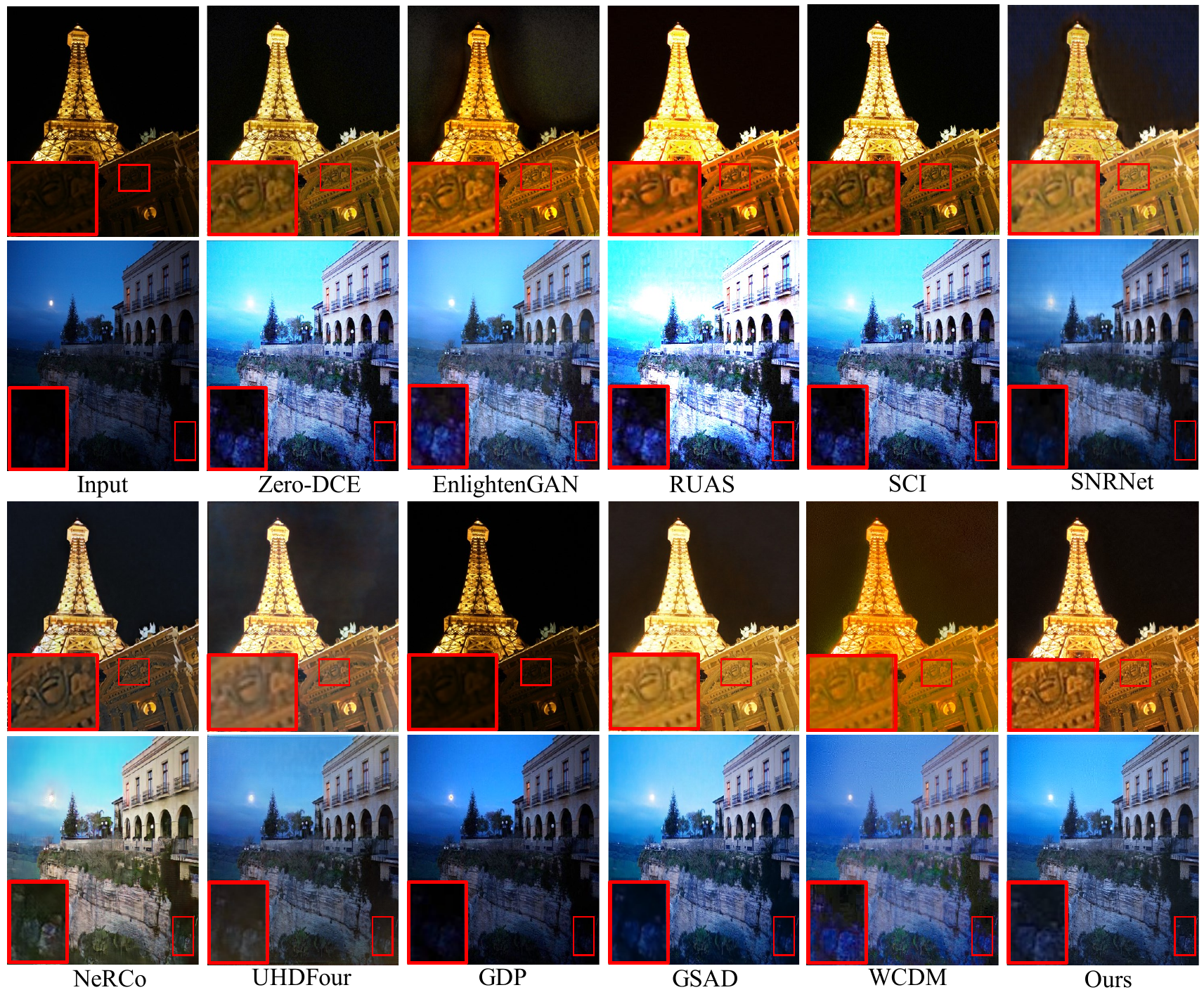}
        \caption{Visual comparison on the DICM \cite{lee2013contrast} (row 1), LIME \cite{guo2016lime} (row 2) datasets among State-of-the-art low-light image enhancement approaches. }
        \label{fig:8}
        \vspace{-10pt}
    \end{figure*}

Meanwhile, we performed evaluation comparisons with competing methods on two unpaired datasets LIME \cite{guo2016lime} and DICM \cite{lee2013contrast} to validate the effectiveness and generalization of our method.  We evaluated the effectiveness of our method in terms of visual quality by combining three non-reference perceptual metrics, NIQE, BRISQUE and PI, with lower metrics resulting in better visual quality. As shown in Table \ref{tab:3}, our method meets the quantification criteria on both datasets compared to other competing methods. Specifically, we obtain the best performance for all quantitative assessments for both NIQE and BRISQUE, while for the PI metrics, we also have the second-best results. This further demonstrates the better generalisability of our approach in real-world scenarios and enhancements that are more in line with human visual perception.

{\bf Visual Comparison.}  Fig. \ref{fig:6} is shown to compare our method with State-of-the-art methods on the paired dataset.  The images in rows 1-3 are selected from the LOLv1, LOLv2-Real\_captured, and LSRW test sets. The visualization of the BAID dataset is then shown in Fig. \ref{fig:7}. Through these comparisons, it is easy to see that previous methods seem to suffer from incorrect exposure, color distortion, noise amplification, or artefacts, which affect the overall visual quality. For example, EnlightenGAN and GDP suffer from generation artefacts and noise amplification, while SNRNet and WCDM suffer from color distortion. In addition GSAD fails to produce similar colours and contrast as the reference image. In contrast, our method consistently produces visually pleasing results with improved color and brightness without overexposure or underexposure. We attribute this to the improved appearance of the multilevel visual-language guidance network. At the same time, CFWD effectively improves contrast, reconstructs sharper details, and brings the visual effect closer to the original image due to the effective constraints imposed by the high-frequency perceptual module on the content structure.



The visual presentation of the DICM and LIME datasets is shown in Fig. \ref{fig:8}. It is clear that our model skilfully adjusts the illumination conditions to optimally improve the contrast of the degraded images while vigilantly avoiding overexposure. This successful balance confirms the generalisability of our proposed method to unseen scenes as well as the satisfaction with the visual results.

\subsection{Ablation Study}
To verify the validity of the proposed method, in this subsection, we will conduct an ablation study of the multiscale visual-language guidance network and the high-frequency perception module, and explore the optimal parameter pairing of the network. All the ablation studies are performed entirely on the LOLv1 dataset.
    
{\bf Multiscale visual-language Guidance Network.} Benefitting from the efficient visual-language prior to CLIP, our method can learn different modalities and thus produce better perceptual and metric results. In order to investigate the effect of the level M of the visual-language guidance network on our method, we fixed the number of wavelet transforms to 2 and verified its effectiveness by gradually increasing the level of visual-language guiding. As shown in Table \ref{tab:4}, when M=0, it indicates that we give up the multimodal learning, and by comparison, we find that after multimodal visual-language guiding, we effectively improve the performance of the network. Meanwhile, with the gradual increase of M, the performance of the network steadily improves. This indicates that multilevel visual-language guidance can iteratively guide the fine-grained alignment of image features with text features during the enhancement process and bring significant network performance improvement.

\begin{table}[h]
\renewcommand\arraystretch{1.4}
\caption{Results of an ablation study at the prompt network scale.}
\scalebox{1.2}{
\begin{tabular}{c|cccc}
\hline
Prompts Scale & PSNR↑           & SSIM↑          & LPIPS↓         & FID↓            \\ \hline
M=0             & 26.705          & 0.856          & 0.227          & 49.926         \\ \hline
M=1             & 27.809          & 0.866          & 0.225          & 48.501          \\ \hline
M=2             & 28.512          & 0.871          & 0.216          & 43.167          \\ \hline
M=3             & \textbf{29.212} & \textbf{0.872} & \textbf{0.197} & \textbf{40.987} \\ \hline
\end{tabular}
}
\label{tab:4}
\end{table}

{\bf Hybrid Frequency Domain Perception Module.} Due to the obvious differences in the feature information contained in the frequency domain space at different stages, we tested a series of combined experiments on the high-frequency perception module, resulting in three HFPM versions. Specifically, HFPM\_v1 uses the wavelet low-frequency domain for Fourier transform to capture image features, HFPM\_v2 uses only the high-frequency space of the first wavelet transform to construct a mixed-frequency domain to capture image information, and HFPM\_v3 performs Fourier transforms on all the wavelet high-frequency domains obtained from multiple wavelet transforms to form a multi-group mixed frequency domain space. By combining multiple sets of mixed-frequency domain spaces, it can effectively acquire high-frequency features. As shown in the Table \ref{tab:5}, the performance of the network using the HFPM\_v1 version is the worst, which may be due to the fact that the wavelet low-frequency domain contains more structural information, which causes more content loss and feature interference when performing the Fourier transform, resulting in a more chaotic feature learned by the model. In addition, compared with HFPM\_v2, HFPM\_v3 has better quantitative results, for the wavelet high-frequency domain, we only need the contour and detail information of the image, therefore, with the combination of multi-group mixing space constraints, we can obtain more detail information to constrain the diversity of the diffusion model content.  

\begin{table}[h]
\caption{Ablation studies of the  optimal effectiveness of our Hybrid Frequency Domain Perception Module.}
\renewcommand\arraystretch{1.4}
\scalebox{1.2}{
\begin{tabular}{l|cccc}
\hline
Versions  & PSNR↑                         & SSIM↑                        & LPIPS↓                       & FID↓                          \\ \hline
HFPM\_v1 & 27.638                        & 0.862                        & 0.215                        & 43.193                        \\ \hline
HFPM\_v2 & {\color[HTML]{333333} 28.282} & {\color[HTML]{333333} 0.868} & {\color[HTML]{333333} 0.209} & {\color[HTML]{333333} 41.185} \\ \hline
HFPM\_v3 & \textbf{29.212}               & \textbf{0.872}               & \textbf{0.197}               & \textbf{40.987}               \\ \hline
\end{tabular}
}
\label{tab:5}
\vspace{-10pt}
\end{table}
\subsection{Discussion}
Despite the excellent performance and visual perception of our proposed low-light image enhancement method, the method still has some non-negligible limitations and goals that need to be further explored. Firstly, the wavelet diffusion model-based low-light enhancement method still has a large computational overhead, which is not conducive to realistic deployment. Second, multiscale visual-language guidance increases the complexity of prompt text design and also carries the risk of augmenting redundant content to some extent. Finally, the loss function required for the enhancement process is more complex, making it difficult to seek the optimal set of weighting parameters. 

In the future, we will investigate a more effective diffusion framework based on the above issues and formulate a more model-compliant visual-language learning network to formulate the appropriate visual-language prompts and remove the risk of redundant content. In addition, the further compact design of the loss function will be the core of our exploration, and through the corresponding research, we believe that the proposed method has further performance space.

\section{Conclusions} \label{Conclusions}
We first successfully introduce multimodal into a diffusion model-based approach for low-light image enhancement and propose a wavelet diffusion model based on CLIP and Fourier transform guidance. By combining the generative power of the diffusion model and the visual-language prior to driving the degraded images for appearance restoration, the visual perception and metric performance are significantly enhanced. In addition, we design a novel high-frequency perception module that effectively constrains the diversity of diffusion model-generated content by exploring the advantages of combining the wavelet and Fourier transforms for double transformation, constructing a hybrid frequency-domain space that is acutely aware of the image structure and provides guidance similar to the target result. Extensive experiments conducted on publicly available benchmark datasets show that our method has better stability and generalisability to provide enhancement of degraded images that approximate the reference image.


\bibliographystyle{IEEEtran}
\bibliography{mybibfile}

\vfill

\end{document}